\documentclass[10pt,twocolumn,letterpaper]{article}
\PassOptionsToPackage{table}{xcolor}
\usepackage{cvpr}              

\usepackage[dvipsnames]{xcolor}

\usepackage{graphicx}
\usepackage{amsmath}
\usepackage{amssymb}
\usepackage{booktabs}
\usepackage{pifont}
\usepackage{color}
\usepackage[linesnumbered,ruled,vlined]{algorithm2e}
\def\degree{${}^{\circ}$}
\def\datasetbig{CSRD-\uppercase\expandafter{\romannumeral5}}
\def\dataset{CSRD-\uppercase\expandafter{\romannumeral2}}

\usepackage{booktabs}       
\usepackage{amsfonts}       
\usepackage{xcolor}         
\usepackage{amsmath}
\usepackage{amssymb}

\usepackage{graphicx}
\usepackage{epsfig}
\usepackage{epstopdf}
\usepackage{makecell}

\usepackage{color}
\usepackage{multirow}
\usepackage{setspace}
\usepackage{enumerate}
\def\degree{${}^{\circ}$}
\usepackage{pifont}

\usepackage[accsupp]{axessibility}
\definecolor{light-gray}{gray}{0.96}

\definecolor{cvprblue}{rgb}{0.21,0.49,0.74}
\usepackage[pagebackref,breaklinks,colorlinks,citecolor=cvprblue]{hyperref}

\def\paperID{2072} 
\def\confName{CVPR}
\def\confYear{2024}

\title{From a Bird's Eye View to See: Joint Camera and Subject Registration \\  without the Camera Calibration}

\author{Zekun Qian\textsuperscript{1}, Ruize Han\textsuperscript{2,3}\footnotemark[2] , Wei Feng\textsuperscript{1}, Song Wang\textsuperscript{4}\\
\textsuperscript{1}College of Intelligence and Computing, Tianjin University\\
\textsuperscript{2}Shenzhen Institute of Advanced Technology, Chinese Academy of Sciences\\
\textsuperscript{3}City University of Hong Kong 
\textsuperscript{4}University of South Carolina\\
{\tt\small \{clarkqian, han\_ruize, wfeng\}@tju.edu.cn, songwang@cec.sc.edu}
}
\begin{document}
\maketitle
\renewcommand{\thefootnote}{\fnsymbol{footnote}}
\footnotetext[2]{Corresponding author.}
\renewcommand{\thefootnote}{\arabic{footnote}}
\begin{abstract}
	We tackle a new problem of multi-view camera and subject registration in the bird's eye view (BEV) without pre-given camera calibration, which promotes the multi-view subject registration problem to a new calibration-free stage. This greatly alleviates the limitation in many practical applications.
	However, this is a very challenging problem since its only input is several RGB images from different first-person views (FPVs), without the BEV image and the calibration of the FPVs, while the output is a unified plane {aggregated from all views} with the positions and orientations of both the subjects and cameras in a BEV.
	For this purpose, we propose an end-to-end framework solving {camera and subject registration together by taking advantage of their mutual dependence}, whose main idea is as below: 
	i) creating a subject view-transform module (VTM) to project each pedestrian from FPV to a virtual BEV, ii) deriving a multi-view geometry-based spatial alignment module (SAM) to estimate the relative camera pose in a unified BEV, iii) selecting and refining the subject and camera registration results within the unified BEV.
	We collect a new large-scale synthetic dataset with rich annotations for training and evaluation. {Additionally, we also collect a real dataset for cross-domain evaluation.} 
	The experimental results show the remarkable effectiveness of our method. 
        The code and proposed datasets are available at \href{https://github.com/zekunqian/BEVSee}{BEVSee}.
\end{abstract}

\vspace{-10pt}
\section{Introduction}
\label{sec:intro}
\begin{figure}[t!] 
	\centering
	\includegraphics[width=0.95\linewidth]{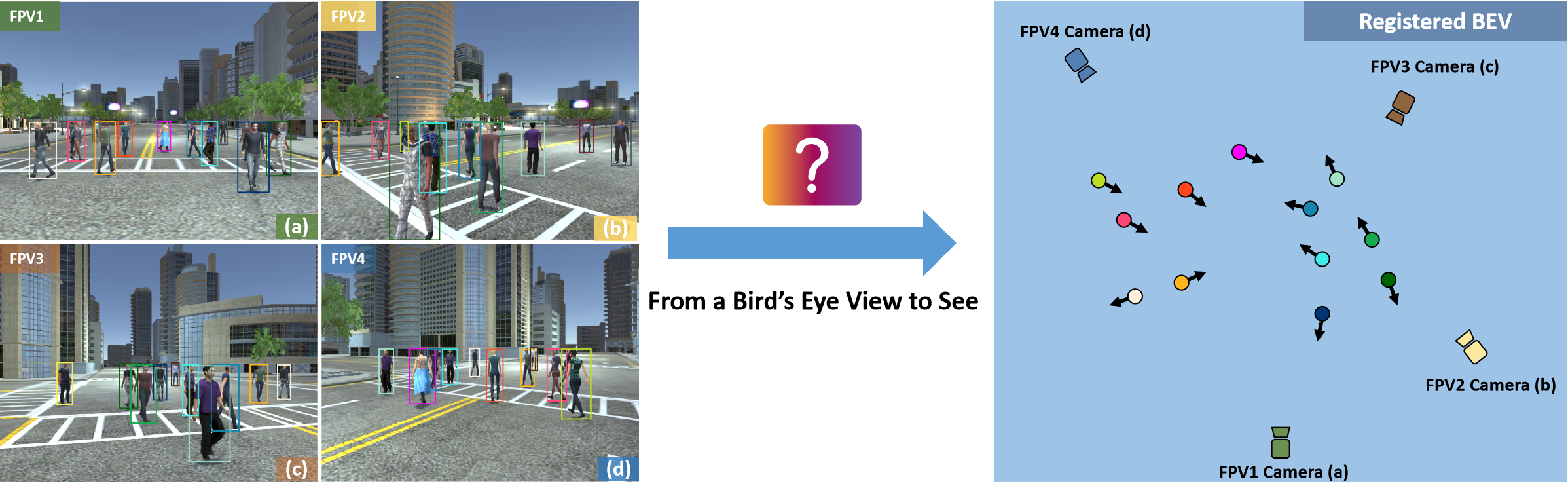}
	\caption{An illustration of the multi-view camera and subject registration problem.}
	\label{fig:question}
	\vspace{-20pt}
\end{figure}

 \textit{There are just three problems in computer vision: registration, registration, and registration.}
 \\ 
 \rightline{-- Takeo Kanade}
Registration is an important task in computer vision. 
In this work, we study a new and challenging problem of \textit{camera and person}  \textit{registration in the BEV without camera calibration}.
Specifically, as shown in Figure~\ref{fig:question}, given the multi-view images for a multi-person scene, {we aim to generate the position and orientation of every person (referred to as subject in this paper) and camera in BEV.} 

This problem is practical for the multi-view camera multi-human scene analysis, which has many applications such as video surveillance, social scene understanding, \textit{etc}.
In this case, the bird's eye view, also called the top view, is a good way to observe the whole scene. In BEV, we can obtain the global spatial layout and trajectories of all the persons in the scene without mutual occlusion, which is very useful in many typical scenarios including the automatic driving~\cite{3d_chen2021monorun,3d_luo2021m3dssd,3d_ma2021delving,3d_wang2021depth,3d_reading2021categorical}, outdoor human detection~\cite{multiview_baque2017deep,multiview_chavdarova2017deep} and complementary-view crowd analysis~\cite{camerahan2020complementary,camerahan2021multiple}.

A popular research problem related to this task is the multi-view human detection \cite{multiview_baque2017deep,multiview_chavdarova2017deep,multiview_hou1_hou2020multiview, multiview_song2021stacked, multiview_hou2021multiview}, which projects the subjects detected from each view to their locations on the ground plane and then generate an occupancy map in the bird's eye view. Note that, these methods all \textit{require the pre-given camera calibration parameters} among the multi-view cameras as input, which, however, limits the applications of the method in many scenes. 
Another series of research focuses on the complementary-view multi-human analysis using a top-view camera (\textit{e.g.}, on a UAV) and several first-person-view cameras \cite{camerahan2019multiple,camerahan2020complementary, camerahan2021multiple,han2022connecting_cvpr}. The main limitation of these methods is that the usage of \textit{a top-view camera carried by a drone is not easy to deploy}.
Similarly, BEV detection in the automatic driving area also relies on the given camera calibration or depth sensor, \textit{e.g.}, LiDAR.

Given the above reasons, in this work, we propose to study a more practical yet challenging problem that achieves not only the subject registration (human localization and face orientation estimation), but also the camera registration (camera localization and view direction estimation) in BEV.
Different from previous works, we register the subjects in BEV but \textit{without} a real bird's-eye-view image, where we generate a virtual BEV. 
Moreover, we do not use the camera calibration as input, but we need to generate a camera registration result, \textit{i.e.}, the camera location and view direction estimation (can be regarded as a weak version of the camera calibration) as output.
This makes this problem very difficult given the \textit{very limited input information and multiple output results}.

A straightforward idea for this problem may be the local descriptors-based methods for multi-view camera pose estimation~\cite{sarlin2020superglue,sun2021loftr,hausler2021patch_match}. However, these algorithms can not handle the proposed problem since they need the input images to have enough overlapped area with textural information. However, in our problem, the overlap may be very limited given the large view difference (even on the opposite side), and the scene (dominant by humans) is unfavorable for local descriptor extraction.
In this work, we propose a novel framework to address this problem. 
Our basic idea is the \textit{registration of the camera and subject are interdependent and complementary}. We \textit{alternately achieve the camera and subject registrations} to make them help each other.
Specifically, the subject distribution in the real 3D world is fixed, which presents variously in the 2D image given different camera poses.
This way, we first restore the subject 3D localization in BEV from the respective camera, and then leverage the prior of the unified subject spatial distribution in the 3D world to estimate the relative camera pose (from subject to camera registration), and finally based on the camera registration to further refine the subject registration in the BEV (from camera to subject registration).

Based on the above insights, we propose a joint framework to simultaneously achieve the subject and camera registration in the BEV. 
Specifically, in each side view, we first apply a view-transform subject detection module (VTM) to obtain the subject detection results in the BEV. 
We then propose a computing-geometry-based spatial alignment module (SAM) to estimate the relative pose of the multiple cameras in the BEV, in which we also apply a self-supervised multi-view human association strategy to obtain the cross-view human corresponding among the multiple views. 
With the camera pose estimation from SAM, in the final registration module, we use a camera pose selection strategy to obtain the camera registration and subject fusion scheme to get unified subject registration in the BEV. 
We summarize the main contributions in this work:\\
\ding{182} To the best of our knowledge, this is the first work to study the camera and subject registration for the multi-view multi-human scene, in which we alternately achieve the camera and human registration results in a unified BEV. This work breaks the limitations of using pre-given camera calibration or real BEV images in previous works. \\
\ding{183} We propose a novel solution for this problem, in which we integrate the deep network-based VTM and a multi-view geometry-based SAM. This framework integrates both the generalization of the deep network for the human localization task and the stability of the classical geometry for the camera pose estimation task.\\
\ding{184} We build a new large-scale synthetic dataset for the proposed problem. Extensive experimental results on this dataset show the superiority of the proposed method and the effectiveness of the key modules. {Furthermore, the cross-domain study on the real dataset verifies the generalization of our method.}

\section{Related Work}
\label{sec:related}

\textbf{Multi-view object detection} seems like the most related work to this paper, which aims to aggregate information about the same object from different views.
The main difficulty of this problem is solving the serious occlusion problem.
Recent approaches for this work are mainly based on the pre-given camera calibration to project the objects detected from each view to
their locations on the real-world ground plane, and then generate an occupancy map in the BEV.
In \cite{multiview_baque2017deep,multiview_chavdarova2017deep}, researchers estimate the positions of pedestrians on the ground plane with corresponding anchor box features.
In \cite{multiview_hou1_hou2020multiview, multiview_song2021stacked, multiview_hou2021multiview}, feature perspective transformation
is employed to project all view features into a shared plane without any anchor.
A couple of datasets have been developed for multi-view pedestrian detection. One is created in a virtual environment, while the other is captured from the real world. These datasets are proposed in~\cite{multiview_chavdarova2018wildtrack, multiview_hou1_hou2020multiview}, respectively.
Besides the multi-view detection, some recent related works have been employed in different fields, \textit{e.g}., cross-view human association and tracking, multi-view 3D human pose estimation~\cite{mutiview_dong2021fast, multivew_tan2019multi, multiview_shere20193d}, \textit{etc}.
Note that, these works all use fixed cameras and require the prior camera calibration as input.
Differently, in this work, we not only do not need the pre-given camera calibration but also provide the camera registration results in the BEV as output.

\begin{figure*}[htpb!] 
	\centering
	\includegraphics[width=0.9\linewidth]
	{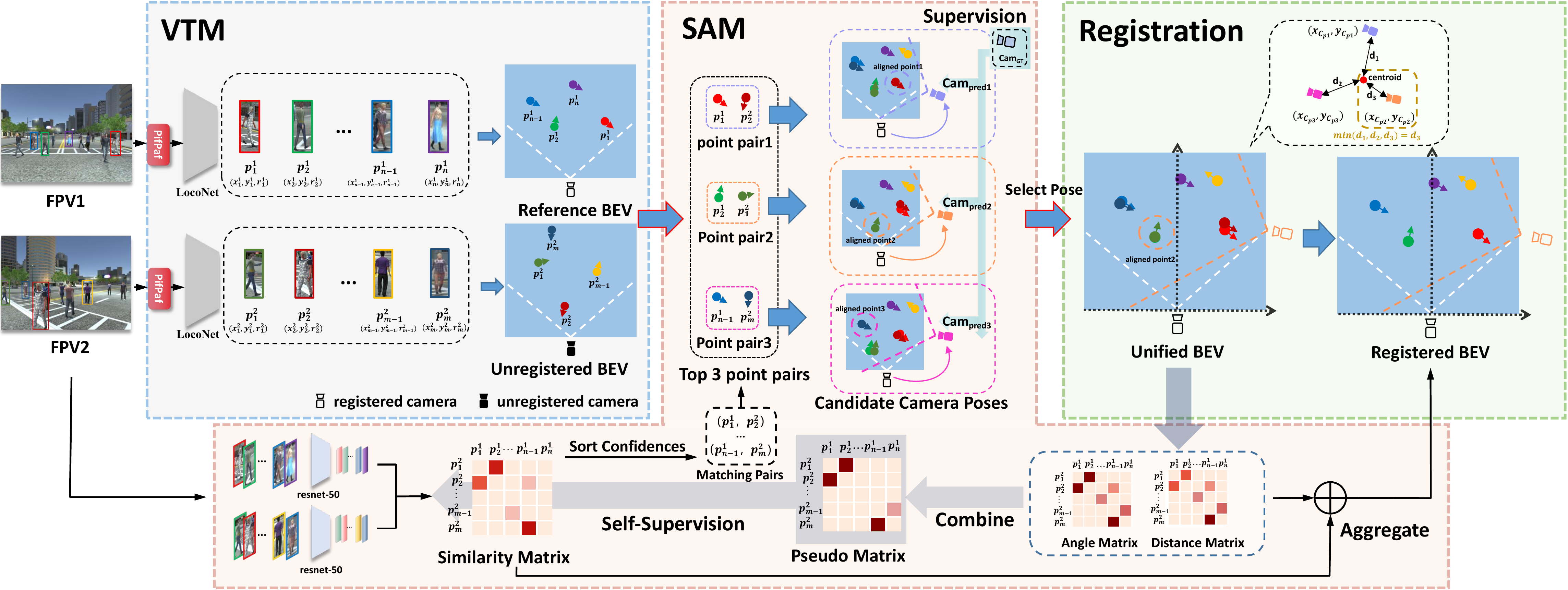}
	\caption{Framework of the proposed method, which can be divided into three parts, \textit{i.e.}, VTM, SAM and Registration. We use hollow camera icons to represent registered cameras and filled camera icons to represent unregistered cameras.}
	\label{fig:framework}
	\vspace{-15pt}
\end{figure*}

{\textbf{3D object detection in autonomous driving} aims to detect objects in traffic scenes. Existing solutions can be broadly classified into three categories. The first category only relies on monocular images, where the localization is directly estimated from monocular images without any depth sensor. 
For instance, general objects are modeled as 3D boxes for localization~\cite{wang2022monocular,li2018stereo,li2022unsupervised,3d_chen2021monorun,3d_wang2021depth,3d_ma2021delving}.
For 3D pedestrian detection works, the 3D skeleton of pedestrians is extracted for localization instead of 3D boxes~\cite{3d_bertoni2019monoloco,monolocopp,3d_hayakawa2020recognition}.
The second category of methods is based on multi-view images. These methods~\cite{wang2022detr3d, chen2022graph, huang2021bevdet, li2022bevformer, jiang2023polarformer,li2023bevdepth} use camera calibration to align different viewpoints and estimate image depth to construct a unified BEV, which can be used to achieve 3D object detection.
The third category of methods~\cite{weng20203d,benbarka2021score,pang2023simpletrack,yin2021center,baser2019fantrack,chaabane2021deft,chiu2021probabilistic,braso2020learning,weng2020gnn3dmot,zaech2022learnable} utilizes depth sensors such as LiDAR to capture the 3D point cloud of the entire scene, based on which they achieve 3D object detection. 
Each of these three categories has its limitations. The monocular view cannot effectively handle occlusions and capture global information. 
The second and third categories rely on the camera calibration and depth sensor, which are costly and have limited applicability.
In contrast to the above mentioned methods, our method not only use the aggregated information from multi-view images but also do not need additional data from camera calibration and depth sensors.}

\textbf{Camera pose estimation} is a related problem to this work, which is a fundamental problem in computer vision. In the long history of its exploration, many methods have been proposed.
Conventional methods to solve this problem used to be helped with some extra measuring devices.
In \cite{cameralaserliu2014external}, the laser rangefinders are used to combine cameras from different views.
In \cite{camerasensordong2016_missing, birdal2016online}, the visual sensors are applied to bridge the huge differences between different fields of views (FOV).
In \cite{cameramobilerobotcensi2013simultaneous,camerarevisitedschonberger2016structure}, some structure from motion (SfM) methods are proposed to track the movement of objects
from different views.
The core for recent vision-based methods~\cite{hausler2021patch_match,sarlin2020superglue} is to find and match the feature points from different views, which, however, are not very useful in this work given the large FOV difference.

\textbf{Bird's-eye-view visual analysis.} Recently, some works have proposed to associate the top view (BEV) with the first-person views for collaborative analysis.
In \cite{camerashi2019spatial,camerashi2020looking}, such idea is employed to locate the first-person-view camera in BEV aerial images in a large field, which is used for GEO-localization.
Later, some related works \cite{camerahan2019multiple,camerahan2020complementary, camerahan2021multiple,han2022connecting_cvpr, han2023relating, han2024benchmarking} focus on the localization of humans,
which aims to associate and track the multiple humans by the spatial reasoning based on the pre-acquired detection.
Another series of works \cite{cameraardeshir2016ego2top, cameraardeshir2018egocentric,cameraardeshir2018integrating} use graph matching-based methods to locate camera wearers by combining information from FPVs and BEV. 
The main difference between the previous works and this work is that they require a BEV image (\textit{e.g.}, captured by a UAV) as input, which is not practical in many real applications.

\section{Proposed Method}
\label{sec:method}

\subsection{Overview}
We first give an overview of the proposed method mainly containing three stages, as shown in Figure~\ref{fig:framework}. 1) Given multiple images simultaneously captured from different views for a multi-human scene, we apply a view-transform subject detection module (VTM) to get the position and the face orientation estimation of each person in the BEV (Section~\ref{sec:vtm}).  2) We then apply a geometric transformation based spatial alignment module (SAM) to estimate the relative camera pose candidates in the BEV (Section~\ref{sec:sam}). 
3) We next use a centroid distance based candidate selection strategy to choose the final camera pose estimation result (camera registration) from the candidates obtained by the SAM. For the subject registration task, we take both spatial and appearance information to aggregate the same person in the BEV for multi-view {subject registration} (Section~\ref{sec:camera_and_subject_registration}).
{Besides, with the subject registration results, we propose a backward training strategy to learn subject similarity for human association in SAM using a self-supervised manner (Section~\ref{sec:self-supervision}).}

\subsection{View-Transform Detection Module (VTM)}
\label{sec:vtm}

For the input of multiple images captured in a multi-human scene, we first get the subject position and face orientation of each person in the BEV.
For this purpose, we develop a \textbf{LocoNet} using a lightweight FC-based structure with three heads. 
\begin{figure}[tpb!] 
	\centering
	\includegraphics[width=0.95\linewidth]{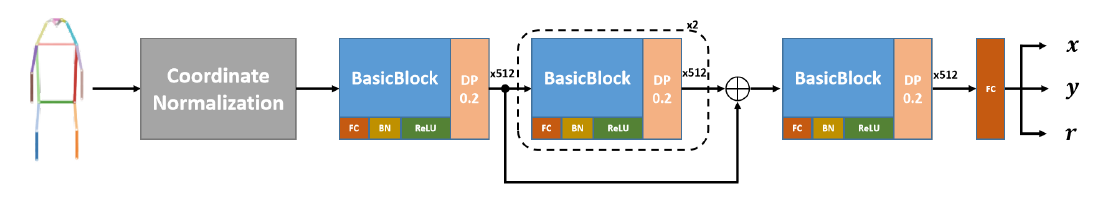} \vspace{-5pt}
	\caption{The structure of LocoNet. 
		Here $\times 512$ means the 512 feature channels, Fc means fully connected layer, BN means batch normalization layer, ReLU is an activation function, DP0.2 is the dropout layer with ratio $0.2$. 
	}
	\label{fig:network}
	\vspace{-10pt}
\end{figure}
Before that, we first apply an existing human pose detector called PifPaf\cite{2019PifPaf} to predict the 2D skeleton joints of each person from the original RGB image, which will be inputted into the LocoNet, whose structure is shown in Figure~\ref{fig:network}.
At the end of {LocoNet}, we use a human 3D localization head composed of simple MLP layers to predict the position and face orientation of each person. 
The process of the network can be represented as 
\begin{equation}
	\label{localnet}
	\begin{aligned}
		\mathbf{p}^{v}_{i} \triangleq (x^{v}_i, y^{v}_i, r^{v}_i)  = \mathrm{LocoNet}(\mathbf{k}^{v}_i),
	\end{aligned}
\end{equation}
where $v$ denotes the $v$-th view and $i$ denotes the $i$-th person in $v$-th view, $\mathbf{k}^{v}_i$ is the 2D skeleton joints belonging to the person $i$ in view $v$. $\mathbf{p}^{v}_{i}$ is the output prediction of {LocoNet}, which is composed of $(x^{v}_i, y^{v}_i)$ representing the subject position in the BEV and $r^{v}_i$ representing the face orientation.

\subsection{Spatial Alignment Module (SAM)} 
\label{sec:sam}
{We then show the relative camera pose estimation (in the BEV) via the subject localization alignment.}
For convenience, we first present the case of two views. 
Our basic idea is that the human position and face orientation are unique in the real-world 3D coordinate system, which can be used for aligning the cameras to generate multiple 2D images.
In the BEV maps with human position and face orientation generated from different first-person view (FPV) images, we can obtain the camera pose in the BEV by aligning the corresponding human position and facing orientation {(as aligned points)} as shown in SAM of Figure~\ref{fig:framework}.

For this purpose, the first step is to find the same subject from different views.
We identify the subjects in the input images through the human appearance features, and the corresponding subjects in the BEV are then matched across different views.
We use a ResNet-50 network to extract the feature of every person and apply Euclidean distance and sigmoid function to create a similarity matrix ($\mathbf{M}_\mathrm{pred}$), which indicates the subject similarities among the subjects from two views.
Then we sort the similarities of each subject pair and select the top-$K$ pairs as the \textit{matching pairs}.

\begin{figure}[htpb!] \vspace{-0pt}
	\centering 
	\includegraphics[width=0.8\linewidth]{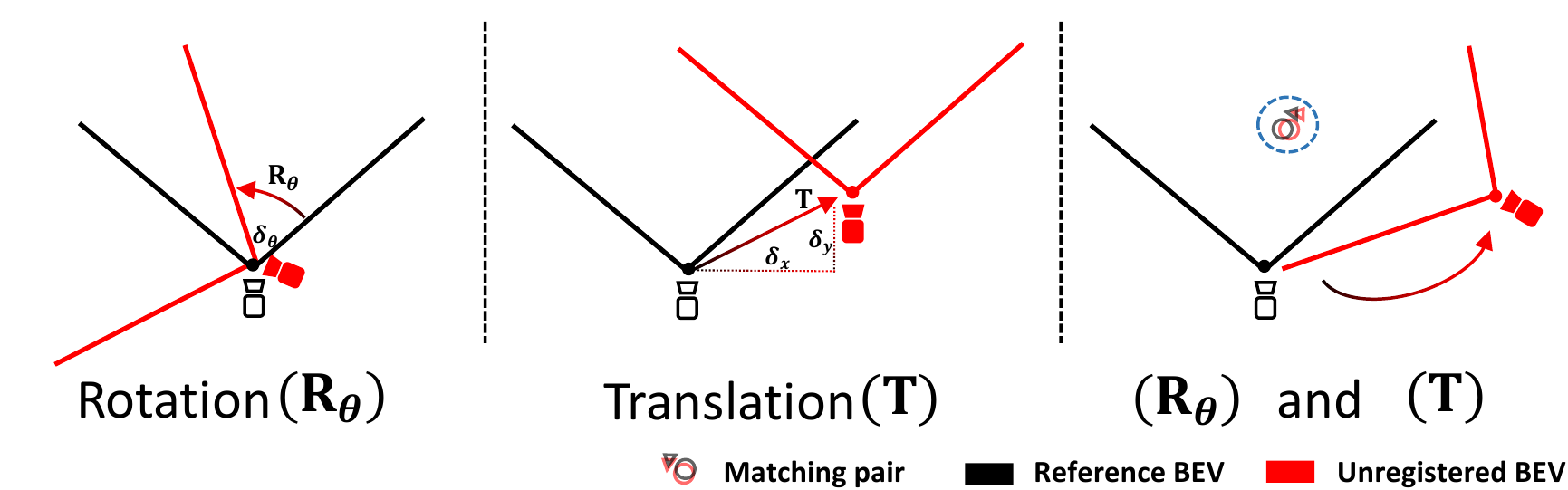}
        \vspace{-5pt}
	\caption{An illustration of the rotation and translation transformation for two BEVs with a matching pair.}
	\label{fig:transformation matrix} 
	\vspace{-15pt}
\end{figure}

After that, we apply the geometric transformation to align two BEVs (containing all subjects and cameras on them), which are denoted as a reference BEV map and an unregistered one.
Specifically, for a pair of matching points, we apply a geometric transformation, as shown in Figure~\ref{fig:transformation matrix}, to rotate and move the camera position and orientation in the unregistered BEV to that in the reference BEV. 

We calculate the relative pose (in the BEV) between two points in a matching pair as discussed above, which are denoted as $\textbf{p}_\mathrm{ref} = (x_\mathrm{ref}, y_\mathrm{ref}, r_\mathrm{ref})$ and $\textbf{p}_\mathrm{unr} = (x_\mathrm{unr}, y_\mathrm{unr}, r_\mathrm{unr})$. Specifically, the relative pose transformation between them can be formulated as

\begin{equation}
	\label{rotation_matrix_and_translation_matrix}
	\begin{aligned}
		\begin{pmatrix}
			x_\mathrm{ref} \\
			y_\mathrm{ref} \\
			1
		\end{pmatrix} & =  \mathbf{T}\mathbf{R_{\theta}} \begin{pmatrix}
			x_\mathrm{unr} \\
			y_\mathrm{unr} \\
			1
		\end{pmatrix},\\
		\quad r_\mathrm{ref} & = \delta_{\theta} + r_\mathrm{unr},\\
	\end{aligned}
\end{equation}	
where 		$\mathbf{R_{\theta}} = \begin{pmatrix}
	\cos\delta_{\theta} & -\sin\delta_{\theta} & 0\\
	\sin\delta_{\theta} & \cos\delta_{\theta} &  0\\
	0		   &	0       &	  1
\end{pmatrix}$, $ \mathbf{T} = \begin{pmatrix}
	1 & 0 & \delta_{x}\\
	0 & 1 &  \delta_{y}\\
	0		   &	0       &	  1
\end{pmatrix}$.
We denote $\mathbf{R_{\theta}}$ as the rotation matrix with a rotation angle $\delta_{\theta}$, and
$\mathbf{T}$ is the translation matrix, in which $(\delta_{x}, \delta_{y})$ is the translation vector.

The corresponding matching pair can be aligned after applying this transformation. 
{This means the transformation matrix is just the relative camera pose between the two cameras in the BEV.}
Note that, this relative camera pose only contains three degrees of freedom, \textit{i.e.}, the translation and rotation projected into the BEV plane.
This way, we can obtain the relative camera pose $(\delta_{x}, \delta_{y}, \delta_{\theta})$ by solving the above Eq.~\eqref{rotation_matrix_and_translation_matrix} and get 
\begin{equation}
	\begin{aligned}
		\begin{cases}
			\delta_{x}&= \ x_\mathrm{ref} - x_\mathrm{unr}\cos\delta_{\theta} + y_\mathrm{unr}\sin\delta_{\theta}\\
			\delta_{y}&= \ y_\mathrm{ref} - x_\mathrm{unr}\sin\delta_{\theta} - y_\mathrm{unr}\cos\delta_{\theta}\\
			\delta_\theta&= \ r_\mathrm{ref} - r_\mathrm{unr}
		\end{cases}.\\		
	\end{aligned}
\end{equation}

As discussed above, we use $K$ point pairs to estimate the relative pose, each of which can generate a relative pose estimation result as shown in SAM in Figure~\ref{fig:framework}. We use the camera pose estimation loss function for training {the LocoNet} as
\begin{equation}
	\label{eq:cam_supervision}
	\begin{aligned}
		\mathcal{L}_\mathrm{Cam} = \sum_{k=1}^K(\|(\delta_{x}^k, \delta_{y}^k) - (\delta_x^\mathrm{gt}, \delta_y^\mathrm{gt})\|+
		\|\delta_{\theta}^k - \delta_{\theta}^{\mathrm{gt}}\|),
	\end{aligned}
\end{equation}
where 
$(\delta_{x}^k, \delta_{y}^k, \delta_{\theta}^k)$ 
is the $k$-th candidate relative camera pose estimation generated from the $k$-th point pair,  and $(\delta_x^\mathrm{gt}, \delta_y^\mathrm{gt}, \delta_{\theta}^{\mathrm{gt}})$ is the ground-truth camera pose.
Note that, we apply the supervision on the camera position, \textit{i.e.}, $\delta_{x}, \delta_{y}$, and the view direction, \textit{i.e.}, $\delta_{\theta}$ in our method. However, the camera view direction is very hard to measure and annotate in real-world applications.
{In the experiments, we show that our method is not very sensitive to the supervision of $\delta_{\theta}$.}

\subsection{Camera and Subject Registration}
\label{sec:camera_and_subject_registration}

\begin{figure}[htpb!]  
        \vspace{-10pt}
	\centering 
	\includegraphics[width=0.6\linewidth]{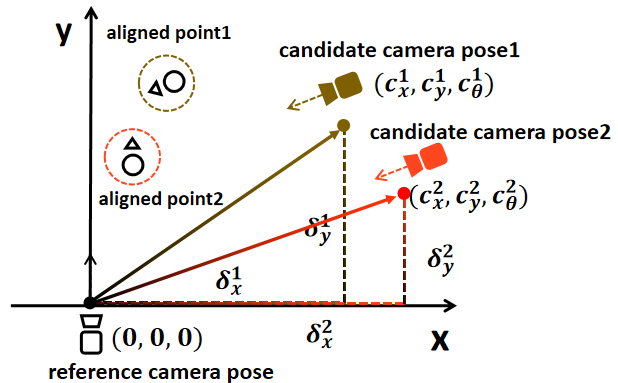}
	\caption{Candidate camera poses in the coordinate system.}
	\label{fig:delta} 
	\vspace{-5pt}
\end{figure}

\textbf{Camera Registration.} Based on the relative camera pose $(\delta_{x}^k, \delta_{y}^k, \delta_{\theta}^k)$ obtained in Section~\ref{sec:sam}, we have got $K$ candidates of relative camera pose estimation between the reference and unregistered BEVs. 
Here we denote the camera pose on the reference BEV as (0, 0, 0). We then get a camera pose of the unregistered BEV  on the coordinate system of the reference BEV as
\begin{equation}
	\textbf{c}^k = (c_x^k, c_y^k, c_{\theta}^k) = (0,0, 0) + (\delta_{x}^k, \delta_{y}^k, \delta_{\theta}^k) =  (\delta_{x}^k, \delta_{y}^k, \delta_{\theta}^k),
\end{equation}
which denotes the $k$-th candidate camera pose from $(\delta_{x}^k, \delta_{y}^k, \delta_{\theta}^k)$, as shown in Figure~\ref{fig:delta}.

The next step is to find the selected camera pose from the $K$ candidates, \textit{i.e.}, achieving the camera registration task. We calculate
$
\overline{c_x} = \frac{\sum_{k=1}^K c_x^k}{K}, \overline{c_y} = \frac{\sum_{k=1}^K c_y^k}{K},
$
where $(c_x^k, c_y^k)$ is the candidate position of the unregistered camera, $(\overline{c_x}, \overline{c_y})$ is the centroid point. 
We then compute the distance of each candidate position to the centroid point as
\begin{equation}
	\label{eq:centroid}
	\begin{aligned}
		{d}^k_\mathrm{centroid} = \|(c_x^k, c_y^k) - (\overline{c_x}, \overline{c_y})\|,
	\end{aligned}
\end{equation}
The candidate with the minimum distance will be selected, which is used to register the unregistered BEV into the reference BEV, then we can get a unified BEV as shown in the left part of Registration in Figure~\ref{fig:framework}.

\textbf{{Subject Registration.}} 
With the camera registration result, we can register the camera position and its view direction, together with the subject localization and face orientation of the unregistered BEV, into the reference BEV.
Note that, for multiple views, we select one as the reference BEV and others as the unregistered BEVs, all of which can be registered into the reference BEV, respectively.
The next step is aggregating the same person from different views in the unified BEV, which can be achieved by two steps, \textit{i.e.}, subject matching and fusion.

1) \textit{Subject Matching.} To match the subjects from multiple views, we create a person spatial distance matrix $\mathbf{M}_\mathrm{dis}$ and an angle difference matrix $\mathbf{M}_\mathrm{ang}$ in the unified BEV, which measure the distance and angle differences of all persons from different views. 
We then combine it with similarity matrix $\mathbf{M}_\mathrm{pred}$ provided in Section~\ref{sec:sam}. 
We first employ three thresholds as filters to select potential matching subject pairs, whereby only pairs that fall within the distance and angle thresholds and surpass the similarity threshold will be identified as the same subject.
Besides, we further consider two constraints for accurate matching.
The first one is cycle consistency~\cite{cycleconsistency}, which means the connection of the same subject from all views should form a loop. 
The second one is uniqueness, which means one subject should not be connected to more than one subject in another view.

For the above constraints, first, we use a classical data structure, \textit{i.e.}, union-find, to aggregate the transitive relations, which makes all the subjects with direct and indirect connections in a {union of} union-find to be clustered as a sub-graph, as shown in Figure~\ref{fig:cutgraph}(b), which solves the problem of cycle consistency for all the subject connected as a loop.
Second, we define the problem as a hierarchical maximum spanning subgraph problem, the layer-by-layer (view-by-view) spanning constrains that a subject is connected at most one node in each view to avoid the uniqueness conflict, as shown in Figure~\ref{fig:cutgraph}(c).
To solve this problem, we propose an algorithm referenced from the Prim algorithm \cite{prim1957shortest}.
We provide more details and the algorithm flow of the above strategy in the \textit{supplementary material}.


\begin{figure}[htpb!] \vspace{-5pt}
	\centering 
	\includegraphics[width=0.95\linewidth]{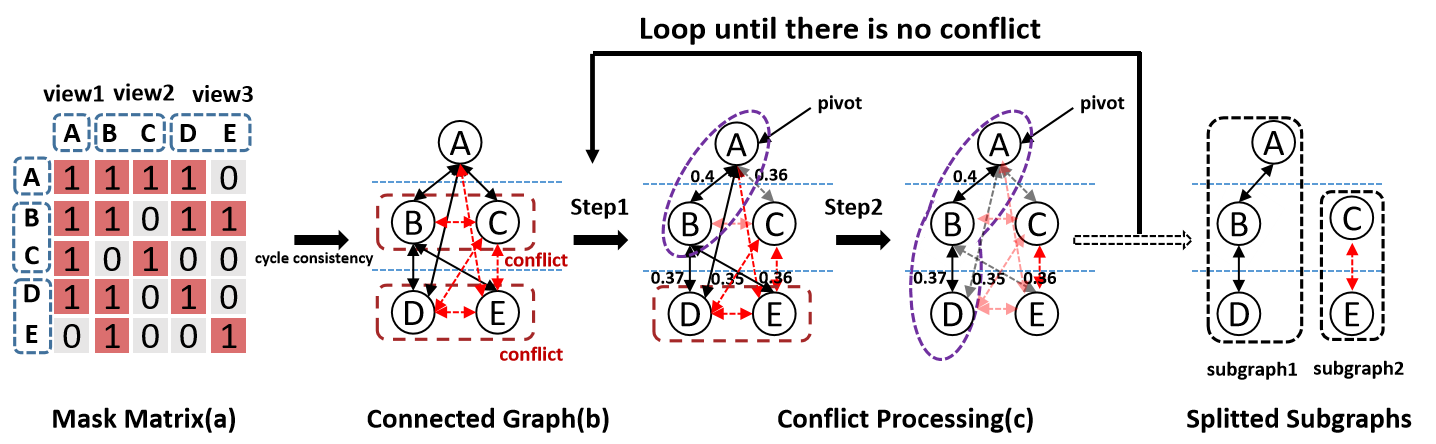}
	\caption{Solving the cycle consistency and uniqueness.}
	\label{fig:cutgraph} 
	\vspace{-10pt}
\end{figure}

2) \textit{Subject Fusion.} For the subjects from multiple views to be regarded as the same person using the above subject matching method, we then estimate {the final registration result of a subject} using the same strategy in Eq.~\eqref{eq:centroid}. 
The position and orientation of the subject with the minimum centroid distance will be retained, and other {same subject} will be removed. 
Especially, if there are only two views, we use the mean position and orientation between two subjects as the fused result.
Finally, we can get the unified BEV with the camera and subject registration from multiple views.

\subsection{Self-supervision for Subject Association}
\label{sec:self-supervision}
Based on the above subject registration results, we further consider to use them for supervising the appearance-based subject association network with a back-propagation strategy. As shown in the bottom of Figure~\ref{fig:framework}, we propose to train the appearance feature extraction network, \eg, { ResNet-50}, for similarity matrix calculating in a self-supervised manner, to make full use of the spatial information from Section~\ref{sec:camera_and_subject_registration}.
Specifically, we inversely normalize each row of the spatial distance matrix $\mathbf{M}_\mathrm{dis}$ and angle difference matrix $\mathbf{M}_\mathrm{ang}$ discussed above, to get the spatial-aware similarity matrixes as
$
\mathbf{M}_\mathrm{spatial} = \alpha \overline{\mathbf{M}_\mathrm{dis}} +  (1 - \alpha)\overline{\mathbf{M}_\mathrm{ang}},
$
where $\alpha$ is the hyper-parameter, $\overline{\mathbf{M}_\mathrm{dis}}$ and the $\overline{\mathbf{M}_\mathrm{ang}}$ are the normalized similarity matrix obtained from  $\mathbf{M}_\mathrm{dis}$ and $\mathbf{M}_\mathrm{ang}$, respectively.
After that, we apply a self-supervised loss to train the appearance feature extraction network as below
\begin{equation}
	\begin{aligned}
		\mathcal{L}_\mathrm{App} = \|\mathbf{M}_\mathrm{pred} - \mathbf{M}_\mathrm{spatial}  \|.
	\end{aligned}
\end{equation}



\subsection{Implementation Details}
\label{sec:implementation_details}
We pretrain the LocoNet using the camera location and view direction labels in our other synthetic data with MSE loss and use the pretrained model of ResNet-50 in \cite{han2022multi_wangyun}.
We use the camera pose loss function and the self-supervised appearance learning loss function as the total loss function
$
\mathcal{L} = \mathcal{L}_\mathrm{Cam} + \mathcal{L}_\mathrm{App}.
$
We set the number of candidate $K$ as 3 in Section \ref{sec:sam} and the similarity matrix threshold as $0.25$, distance threshold as 2.0 $m$ and angle threshold as 15\degree \ in Section~\ref{sec:camera_and_subject_registration}.
We set the hyper-parameters of the pseudo matrix in Section~\ref{sec:self-supervision} as $\alpha = 0.5$. 
{We use a pair of FPVs to train our framework. In the inference stage, the number of FPVs is not limited, in which one FPV will be selected as the reference view and others can be registered in that reference view.}
We use Pytorch as our main framework and the work runs on the server with RTX 3090 GPU.

\vspace{-4pt}
\section{Experiments}
\label{sec:experiment}

\subsection{Proposed Dataset}
\label{sub_sec:dataset}
To our best knowledge, there is no available dataset that can be used for the task in this work, which requires the multi-view relative camera poses, the 3D position and the face orientation of each person. Even with expensive hardware, it is also very hard to obtain accurate annotations of them in the real world.
So we consider using the modeling engine to create a synthetic dataset.\\
$ \bullet $ \textit{Flexible data controlling.} We use a 3D game development Unity 3D~\cite{unity} to build a city scene and apply open-source 3D human model library PersonX~\cite{personx} containing more than 1,000 different persons to generate subjects in the scene. 
Through the flexible development platform, we can create various scenes to simulate the real world. 
The cover area is set as $25m \times 25 m$, in which all the objects are simulated to the real environment with a scaling.
\\
$ \bullet $ \textit{Diverse subject settings.}
For data diversity, the number of subjects in each frame is different, 
where the range of subjects in the scene is from 10 to 25, containing  5-20 people walking free and 5 camera-wears.
Further, we generate every frame by random function, which means camera registration and subject registration are various in each frame.
\\
$ \bullet $ \textit{Large scale.} 
We create two \textit{C}amera \textit{S}ubject \textit{R}egistration \textit{D}atasets, \textit{i.e.}, {\dataset}
and {\datasetbig}, which contain two views and five views, respectively.
In total, {\dataset} includes 2,000 pairs of images, with 1,000 for training and another 1,000 for testing.
{\datasetbig} includes 1,000 groups of images, in which each group contains 5 synchronous images. {\datasetbig} is only used for testing in our experiments.
\\
$ \bullet $ \textit{Rich and accurate annotations.} 
Our annotations contain the position (in meters) and face orientation of each subject in the BEV, as well as the camera poses. 
Besides, we also provide the bounding box with the unified ID number of each subject in all views. More details about the proposed datasets can be obtained in the \textit{supplementary material}.



\subsection{Setup}
\label{sub_sec:setup}
\textbf{Evaluation Metrics.} 
\textbf{Metric-I}: We first evaluate the accuracy of the \textit{camera registration} results, including the position and orientation results in the BEV. 
For the position, we calculate the distance between the predicted and ground-truth positions. Then we count the average error (\textit{Cam.Pos.Avg}) and the percentages of the error within a list of a certain distance, including 0.5, 1, and 1.5 meters. 
Similarly, we calculate the angle error in average (\textit{Cam.Ori.Avg}) and percentages of degree errors within certain ranges, including 5, 10 and 15 degrees.
\textbf{Metric-II}: We also evaluate the \textit{subject registration} results.
It is similar to the metric-I, which evaluates the position distance and orientation error of the subjects. 
\textbf{Metric-III}: We finally evaluate the multi-view multiple human association (identification) results. We use precision, recall, and $F_1$ scores as the metrics.

\begin{table*}[htbp]
	\caption{Camera registration results. The top half is comparison experiments, the bottom half is ablation study, in which `Cam.Pos.Avg' and `Cam.Ori.Avg' present the average error in meters of the camera position and the orientation error in degrees in BEV, `Cam.Pos@$d$' represents the percentage of distance error within $d$ meters and `Cam.Ori.@$r$' represents the percentage of angle error within $r$ degrees.}
	\vspace{-20pt}
	\label{tab:camera}
	\centering
	\rowcolors{2}{}{light-gray}
	\footnotesize
	\tabcolsep=0.17cm
	\begin{spacing}{1.15}
		\begin{center}
			\begin{tabular}{lcccccccc}
				\Xhline{1pt}
				Methods & Cam.Pos.Avg & Cam.Ori.Avg & Cam.Pos@0.5 & Cam.Pos.@1 & Cam.Pos.@1.5 & Cam.Ori.@5 & Cam.Ori.@10 & Cam.Ori.@15 \\ \hline
				Monoloco++~\cite{monolocopp}        & 3.00        & 21.84     & 7.60\%       & 21.60\%    & 36.40\%    & 17.50\%   & 34.60\%    & 47.10\%    \\ \hline
				DMHA~\cite{han2022connecting_cvpr}                & 5.99         & 47.43      & 46.50\%      & 47.60\%    & 48.60\%    & 46.20\%   & 50.00\%    & 53.60\%    \\ \hline
				SIFT~\cite{lowe2004distinctivesift}           & 7.11        & 144.46     & 1.26\%       & 2.34\%     & 3.60\%     & 4.80\%    & 8.20\%     & 11.10\%    \\ \hline
				LoFTR~\cite{sun2021loftr}  &11.50	&90.11 &0.70\% &1.20\% &1.70\% &3.70\% &6.50\% &8.50\% \\ \hline
				SuperGlue~\cite{sarlin2020superglue} &11.17	&89.74 &0.60\% &1.10\% &1.50\% &3.70\% &6.50\% &8.60\% \\ \hline
                CVNet~\cite{lee2022correlation} & 11.38 & 115.10  & 0.88\% & 1.25\% & 1.75\%& 3.10\%& 5.5\%& 7.40\%  \\  \hline
                R2Former~\cite{zhu2023r2former} & 13.55 & 102.52 & 0.35\% & 0.47\% & 0.83\% & 3.90\% & 7.20\%& 9.50\% \\
                
                \hline \hline
				Max                  & 2.27        & 15.22     & 20.00\%      & 42.30\%    & 59.60\%    & 33.90\%   & 60.30\%    & 76.00\%    \\ \hline
				Random               & 1.91        & 12.62     & 21.60\%      & 47.30\%    & 65.00\%    & 37.50\%   & 65.80\%    & 81.20\%    \\ \hline
				w/o pre-train   & 6.98         & 33.02     & 0.50\%       & 1.40\%     & 3.20\%     & 10.20\%   & 20.90\%    & 29.50\%    \\ 
				\hline
				w/o GT $\delta_{\theta}$ & 0.93        & 5.91      & 37.80\%      & 71.80\%    & 85.60\%    & 59.10\%   & 85.60\%    & 94.30\%    \\	
				\hline \hline
				Ours      & 0.89 & 5.78 &42.20\% & 72.40\% & 88.40\% & 59.50\% & 86.50\% & 94.80\%        \\ 
				\Xhline{1pt}
			\end{tabular}
		\end{center}
	\end{spacing} \vspace{-0.5cm}
\end{table*}
\begin{table*}[htbp]
	\caption{Subject registration results. The expression of metrics of subject here is in the same way as Table~\ref{tab:camera}.}
	\vspace{-20pt}
	\label{tab:subject}
	\centering
	\rowcolors{2}{light-gray}{}
	\footnotesize
	\begin{spacing}{1.15}
		\begin{center}
			\begin{tabular}{lcccccccc}
				\Xhline{1pt}
				Methods & Sub.Pos.Avg & Sub.Ori.Avg & Sub.Pos.@0.5 &  Sub.Pos.@1 &  Sub.Pos.@1.5 &Sub.Ori.@5 &Sub.Ori.@10 & Sub.Ori.@15 \\ \hline
				Monoloco++~\cite{monolocopp}    & 1.32           & 32.50        & 26.05\%         & 61.47\%       & 77.65\%       & 13.21\%      & 26.05\%       & 38.17\%       \\ \hline
				MVDetr~\cite{multiview_hou2021multiview}               & 2.41           & -             & 11.18\%         & 29.54\%       & 46.07\%       & -          & -           & -           \\ \hline
				MVDet~\cite{multiview_hou1_hou2020multiview}                & 2.44           & -             & 11.28\%         & 29.19\%       & 45.65\%       & -          & -          & -          \\ \hline \hline
				w/o pre-train   & 6.35           & 89.29       & 1.62\%          & 6.62\%        & 11.41\%       & 2.29\%       & 4.74\%        & 6.97\%        \\ \hline
				w/o GT $\delta_{\theta}$ & 0.83           & 16.36        & 41.15\%         & 77.89\%       & 89.31\%       & 32.30\%      & 56.79\%       & 72.77\%       \\	 \hline
				Max                  & 1.27           & 21.56        & 37.39\%         & 72.38\%       & 82.87\%       & 30.46\%      & 54.95\%       & 69.13\%       \\ \hline
				Random               & 1.06           & 17.19        & 39.19\%         & 74.62\%       & 85.07\%       & 33.61\%      & 59.01\%       & 73.39\%       \\ \hline
				\hline 
				Ours   & 0.75 & 14.67 & 43.23\% & 81.43\% & 92.12\% & 35.07\% & 63.24\% & 79.15\%  \\           
				\Xhline{1pt}
			\end{tabular}
		\end{center}
	\end{spacing} \vspace{-0.8cm}
\end{table*}

\textbf{Comparison Methods.} 
As discussed above, there is no method that can directly handle the proposed problem. 
We include the following comparison methods for the \textit{camera registration} task:  \textit{DMHA}~\cite{han2022connecting_cvpr} achieves the task of camera registration by using a real BEV image.  We provide the FPV images and the corresponding BEV image to DMHA.
\textit{SIFT \cite{lowe2004distinctivesift} + KNN} and {other {deep-learning-based methods} \cite{sun2021loftr,sarlin2020superglue,lee2022correlation,zhu2023r2former} are local descriptor (key point) matching based methods}, which are combined with the classical camera pose estimation method with the matched key points for relative camera estimation.
For the second task of \textit{subject registration}, we include the following three methods.
\textit{Monoloco++} \cite{monolocopp} is a network predict 3D-localization and face orientation of each person in the view. We concatenate it with our geometric transformation and subject fusion methods for comparison.
\textit{MVDet} and \textit{MVDetr} \cite{multiview_hou1_hou2020multiview,multiview_hou2021multiview} are used for multi-view object detection with the camera calibrations. 
We provide more details about the implementation details of the comparison methods in the \textit{supplementary material}.

\subsection{Comparative Results}
\label{sub_sec:camera_result}
\textbf{Camera Registration Results.}
We first evaluate the camera registration results on {\dataset} as shown at the top half in Table~\ref{tab:camera}.
We can first see that all the comparison methods provide very poor results.
Among them, we provide the ground-truth BEV image to DMHA, which is used to find the camera wearer from the BEV instead of our position regression. 
The key point matching based methods almost fail because of the huge view differences.
Monoloco++ generates a relatively acceptable result since it's equipped with the proposed geometric transformation methods.
For our method, the mean distance error is only $0.89$ meters, less than $1$. The most remarkable thing is the accuracy under camera angle error $\leq15$ degrees is more than $94\%$, even $\leq5$ degrees is up to $59\%$, and the mean error is less than $6$ degrees.
{This is promising for many real-world applications.}

\textbf{Subject Registration Results.}
\label{sub_sec:subject_result}
We also evaluate the subject registration in {\dataset} using Metric-II as shown in Table~\ref{tab:subject}. 
Even MVDet and MVDetr take the camera calibration as prior, our method achieves much superior results in all metrics.
{At the same time, our method keeps the average distance error within 0.8 meters and the average orientation error within 15 degrees.} 

\subsection{Ablation Study}
$ \bullet $ w/o pre-train.: Removing the pre-training of LocoNet.\\
$ \bullet $ w/o GT $\delta_{\theta}$: Removing the supervision of the camera orientation in Eq.~\eqref{eq:cam_supervision}.\\
\noindent$ \bullet $	Max/Random: In the candidate camera selection strategy, we choose the max confidence pair or choose randomly instead of our method in Eq.~\eqref{eq:centroid}.

\setcounter{table}{3}
\begin{table*}[htbp!]
\vspace{-20pt}
\caption{Multi-view camera and subject registration, and multi-view subject association results.}
\vspace{-20pt}
\label{tab:multiview}
\centering
\rowcolors{2}{}{light-gray}
\tabcolsep=0.1cm
\footnotesize
\begin{spacing}{1.15}
	\begin{center}
		\begin{tabular}{lccccccccc}
			\Xhline{1pt}
			Methods	&  Cam.Pos.Avg & Cam.Ori.Avg & Cam.Pos.@1  & Cam.Ori.@10  
			&Sub.Pos.Avg & Sub.Ori.Avg &  Sub.Pos.@1 &Sub.Ori.@10   & $F_1$   \\ \hline
			Pair-wise     & 1.06        & 6.96         & 62.71\%     & 79.61\%     & 0.75           & 14.67              & 80.76\%         & 58.81\%       & 83.85\%  \\ \hline 
			Multi-view w/o constraints & 1.06         & 6.93        & 63.55\%   & 80.60\%     & 1.10           & 15.64             & 63.78\%           & 50.47\%    	& 85.64\%    \\ \hline 
			Multi-view w constraints     & 1.06        & 6.93         & 63.55\%     & 80.60\%     & 0.94           & 13.45              & 70.57\%         & 57.73\%       & 86.12\%  \\ 	\Xhline{1pt} 
		\end{tabular}
	\end{center}
\end{spacing} \vspace{-0.8cm}
\end{table*}

As shown at the bottom half of Table~\ref{tab:camera}, the ablation study, \textit{i.e.}, `	w/o pre-train', verifies the necessity of the pretrained LocoNet in VTM. 
{We can also see from the next row that, when removing the camera orientation supervision in SAM, \textit{i.e.}, `w/o GT $\delta_{\theta}$', the performance only drops a little. This demonstrates that our method \textit{is not heavily dependent on} the camera orientation supervision, which is not easy to obtain in the real world.}
For camera pose selection in Registration module, we can see that no matter whether using the strategy of the max confidence one or the random one, which, not considering the spatial-aware selecting strategy, both provide a relatively poor performance than our centroid strategy. 
We also conduct the ablation study on the subject registration task, as shown in Table~\ref{tab:subject}. Similar to the above results, we can see the effectiveness of the pretrained LocoNet in VTM, camera orientation supervision in SAM, and the centroid strategy in Registration module.

We further evaluate the results of multi-view human association in {\dataset}, which can verify \textit{the effectiveness of the proposed backward self-supervised training strategy in SAM}. As shown in Table~\ref{tab:re-id}, the baseline is the ResNet-50 model pre-trained on the person Re-ID dataset named Market-1501~\cite{zheng2015scalable_market}, on which we apply the self-supervised training strategy as discussed in Section~\ref{sec:self-supervision}. `w GT re-id' denotes that we provide the ground-truth assignment matrix to supervise the result of the similarity matrix.
We can see from Table~\ref{tab:re-id} that our self-supervision strategy improves the $F_1$ score from the original $66.78\%$ to $85.98\%$ with a large margin. We can also see that our results are very close to the result of supervised training, with only a small gap of $0.45\%$. 
The results of the association verify the effectiveness of the proposed self-supervision strategy.

\setcounter{table}{2}

\begin{table}[htbp!]
\vspace{-10pt}
\caption{Cross-view subject association results.}
\vspace{-20pt}
\label{tab:re-id}
\rowcolors{2}{}{light-gray}
\centering
\footnotesize
\begin{spacing}{1.25}
	\begin{center}
		\begin{tabular}{lccc}
			\Xhline{1pt}
			Methods & Precision  & Recall         & $F_1$     \\ \hline
			Baseline~\cite{han2022multi_wangyun}                   &57.48\%    &82.98\% & 66.78\%      \\ \hline
			Ours                       &79.33\%    &95.45\%   & 85.98\%      \\ \hline
			w GT re-id (oracle)        &77.97\%     & 98.18\%  & 86.43\%    \\ 
			\Xhline{1pt}
		\end{tabular}
	\end{center}
\end{spacing} \vspace{-0.3cm}
\end{table}
\vspace{-10pt}

We further evaluate the proposed method on the scenes using multiple cameras for camera and subject registration on {\datasetbig}, as shown in Table~\ref{tab:multiview}.
The first row shows the results that we split the 5 views into $C^{2}_{5}$ = 10 pair-wise views, on which we apply the proposed method for two views as above.
The second and third rows are the results of multi-view subject and camera registration without or with the constraints during subject matching in Section~\ref{sec:camera_and_subject_registration}. 
We can see that using the proposed constraints effectively improves the results on subject registration and multi-view subject association, which demonstrates \textit{the effectiveness of the subject registration strategy in the Registration module}.
With respect to the results of two views registration, we can see that even though the results are slightly worse in 5 views, the overall results are still very impressive, which demonstrates the stability of our method in multiple views.

\subsection{In-depth Analysis}
\label{sub_sec:others}

\textbf{Real-world Dataset Evaluation.}
\label{sub_sec:real}
We propose a large-scale real-world \textit{evaluation  dataset} {CSRD-R}, to test the performance of the cross-domain of our method, which includes 15,490 frames and five different scenes. There are 1,500 synchronous frame groups for the two-view scene, 830 synchronous frame groups for the three-view scene, and 2,500 synchronous frame groups for the four-view scene.
In addition to the first-person views provided by the wearing cameras, we also capture the real BEV using a UAV.
For all the  first-person-view and BEV videos in the dataset, we annotate the bounding boxes for each subject and label the unified ID for same subject in all views.
Next, we conducted cross-domain experiments where we train our model on the synthetic dataset {\dataset} and performed a cross-domain evaluation on the real dataset, CSRD-R.




As shown in Table~\ref{tab:real cos}, we provide the detection performance of our method. Note that, considering the gap between the BEV generated by our method and the real BEV, we define the new detection metric on CSRD-R, which is provided in the \textit{supplementary material}. The results demonstrate the effectiveness of our method on real data and its reliable cross-domain generalization ability.

\setcounter{table}{4}

\begin{table}[htbp]
\vspace{-0pt}
\caption{Results on CSRD-R for different numbers of views.}
\vspace{-15pt}
\label{tab:real cos}
\centering
\rowcolors{2}{}{light-gray}
\footnotesize
\begin{spacing}{1.15}
	\begin{center}
		\begin{tabular}{lccc}
			\Xhline{1pt}
			& Two Views & Three Views & Four Views \\ \hline
			Ours & 82.50\% & 85.07\% & 86.31\% \\ \hline
			\Xhline{1pt}
		\end{tabular}
	\end{center}
\end{spacing} 
\vspace{-20pt}
\end{table}

\begin{figure}[htpb!] \vspace{-5pt}
\centering
\includegraphics[width=0.9\linewidth]{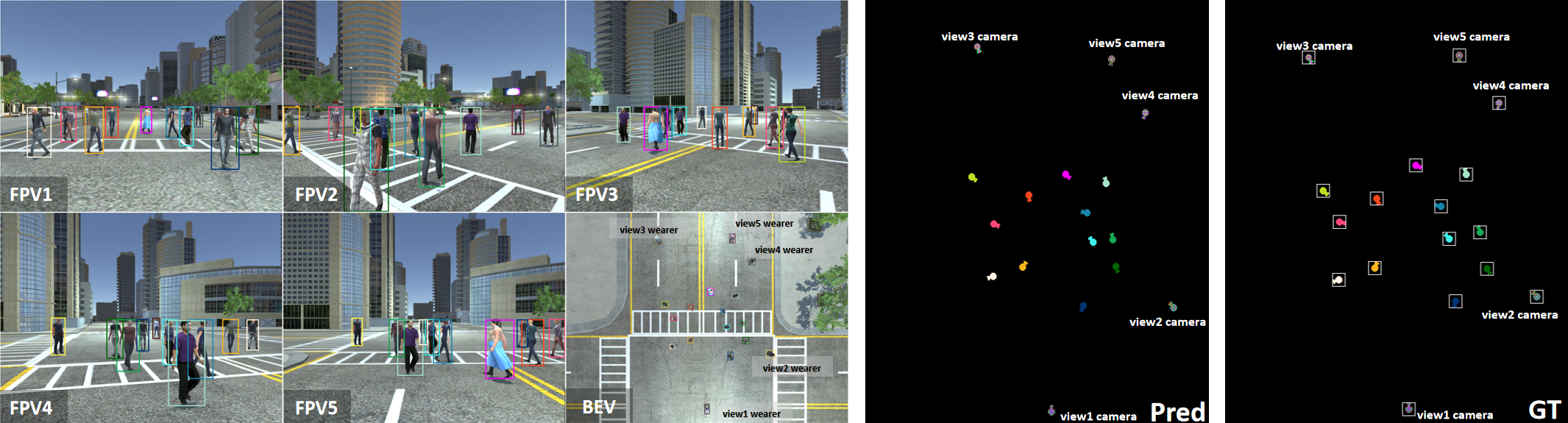} 
 \vspace{-5pt}
\caption{Qualitative case analysis. We add a white rectangle around every ground-truth subject.} 
\label{fig:great_case1}
\vspace{-10pt}
\end{figure}

\begin{figure}[htpb!] \vspace{-5pt}
\centering
\includegraphics[width=0.9\linewidth]{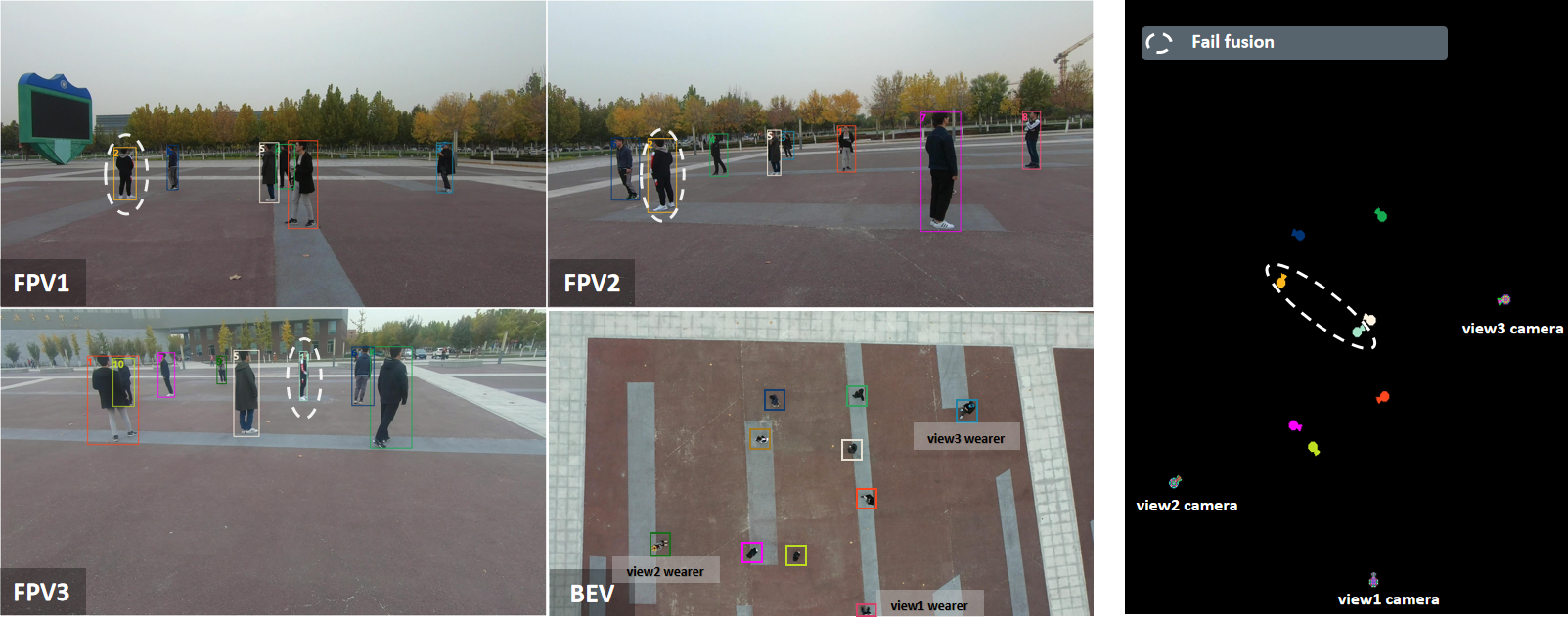} 
 \vspace{-5pt}
\caption{Real-world case study.} 
\label{fig:real_case1}
\vspace{-15pt}
\end{figure}

\textbf{Qualitative Analysis.}
Figure~\ref{fig:great_case1} 
shows a case, in which we can see that the prediction of both camera and subject registration can achieve a good coincidence with the ground truth, thanks to the high accuracy of our method.
We also provide the real-world case, as shown in Figure~\ref{fig:real_case1}. 
Note that, we directly apply our method to the real-world case without any additional annotation.
We can see that, except for a wrong fusion coming from the incorrect matching, the prediction of the subject and camera distributions are very close to the real BEV. This demonstrates the robustness and generalization of the proposed method.
More visualization results with special conditions and analyses on sensitivity and complexity are available in the \textit{supplementary material}.


\section{Conclusion}
\label{sec:conclusion}
In this paper, we have studied a new problem of multi-view camera and subject registration tasks in BEV without camera calibrations.
For this problem, we develop a new approach that can simultaneously handle these two tasks.
Specifically, the proposed method uses an end-to-end framework, which makes full use of deep network based appearance information and multiple view geometry based spatial knowledge to complement each other's advantages.
We also create new synthetic and real-world datasets with various settings and rich annotations. 
Experimental results show the superior performance of our method.

\textbf{Acknowledgment.} This work was supported in part by the NSFC under Grants 62072334, U1803264.

\section{Supplementary Material}
\subsection{Details of the Subject Matching Algorithm}

As presented in the `\textbf{Subject Registration}' in Section 3.4. 
We consider two constraints for accurate matching. The first
one is cycle consistency, which means the connection
of the same subject from all views should form a loop. 
The second one is uniqueness, which means one subject should
not be connected to more than one subject in another view.
To clearly explain the two constraints and our solutions. We present an example to illustrate.

After applying the binarization operation with thresholds, we can get a mask matrix to show which pairs may be the same person, as illustrated in Figure~\ref{fig:cutgraph}(a). Analyzing the matrix, we can know that person A in view 1 matches both persons B and C in view 2 and person D in view 3. Similarly, person B in view 2 may match person D and E in view 3. The black arrows in Figure~\ref{fig:cutgraph}(b) visually represent the matching relationships. 
\begin{figure*}[h] 
	\centering 
	\includegraphics[width=0.95\linewidth]{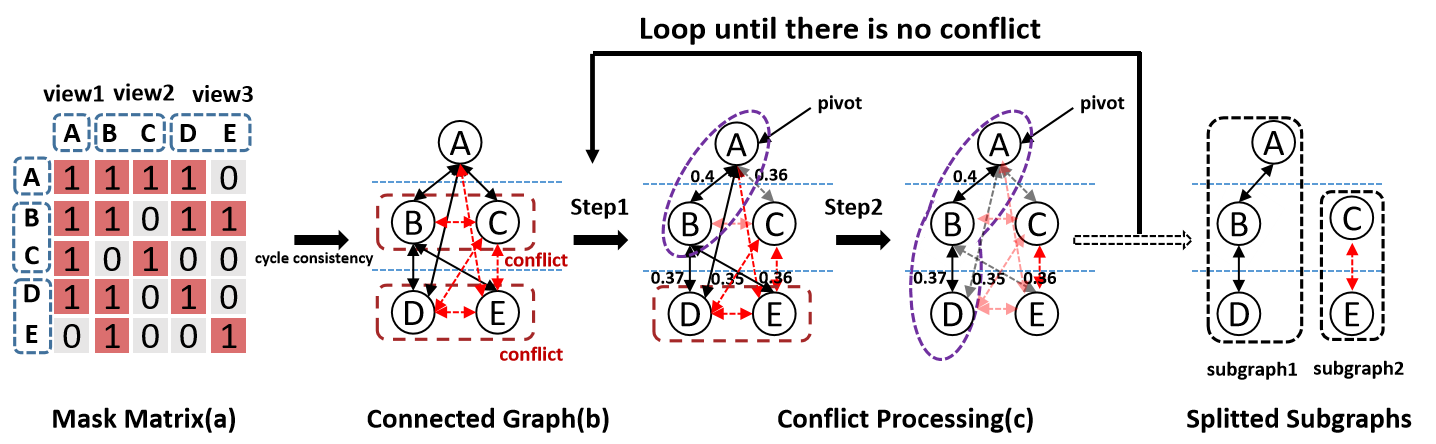}
	\caption{An example of solving the cycle consistency and uniqueness.}
	\label{fig:cutgraph}
\end{figure*}

Considering the matching results, the first problem here is the lack of cycle consistency. 
we can see that A and B are connected, as well as B and E are connected. If these two connections are correct, the cycle consistency requires that A and E should also be connected as the same person. 
But we can see from the mask matrix that A and E have no connection between them.

\begin{algorithm}[htbp]	
	\small
	\caption{Uniqueness conflict solving:}
	\label{alg:split_complete_graph}
	\KwIn{${S}_\mathrm{ori}$: A set of node indices with some uniqueness conflicts, \\\quad\quad\quad $M_\mathrm{pred}$: A similarity score matrix between all persons, \\\quad\quad\quad${\rm Mask}$: A mask matrix to denote the connection between different persons.}
	\KwOut{${L}_\mathrm{res}$: A list of sets of node indices without uniqueness conflict.}
	${L}_\mathrm{res}$ = [] //used to record the result of divided subgraphs.
	
	${L}_{view}$ = DivideGraphByView(${S}_\mathrm{ori}$) //Dividing the nodes into different views.
	
	\While{\rm{Length}(${L}_\mathrm{view}$) $> 0$}{
		
		n = Length(${L}_\mathrm{view}$)
		
		pivot = ${L}_\mathrm{view}$[0][0]
		
		tmp\_set = new set()
		
		tmp\_set.add(pivot)
		
		${L}_\mathrm{view}$[0].pop(pivot) // Removing the pivot node from the original graph.

		\For{$v = 1 : n-1$}{
			
			node = GetMaxScoreOfView(${L}_\mathrm{view}$[$v$], tmp\_set, $M_{pred}$, ${\rm Mask}$) //Used to get the selected node in this view, if no node meets the condition will return -1.
			
			\If{${\rm node}\quad! = - 1$}{
				tmp\_set.add(node)//Adding the selected node to subgraph.
				
				${L}_\mathrm{view}$[v].pop(node)//Removing the selected node from the original graph.
			} 
		}
		${L}_\mathrm{res}$.append(tmp\_set)//Saving the subgraph.
		
		RemoveEmptyView(${L}_\mathrm{view}$)//Removing the layer(view) with no node remaining.		
		
	}
	\Return ${L}_\mathrm{res}$
\end{algorithm} 

To solve the problem, we use a data structure called union-find to aggregate the transitive relation in the mask matrix. For every aggregated union from the union-find, we create an augmented graph with hidden edges as the red dashed arrow shown in Figure~\ref{fig:cutgraph}(b).
Now, there is an implicit connection between A and E by the transitive path: A to B and B to E, where the weight of each edge is the confidence score from the similarity matrix.

We divide the nodes of the graph into different layers (representing different views) as separated by blue dashed lines in Figure~\ref{fig:cutgraph}(b).
We can find some conflicts of uniqueness between B and C (both connected to A), D and E (both connected to B) in the graph, as highlighted within red dashed rectangles. 
We consider cutting the graph into reasonable subgraphs without uniqueness conflicts. 
We define the problem as a hierarchical maximum spanning sub-graph problem, the layer-by-layer (view-by-view) spanning constraint that a subject is connected at most to one node in each view to avoid the uniqueness conflict.
Figure~\ref{fig:cutgraph}(c) shows the complete flow of our solution of an example. 

Specifically, first, we select A from view 1 as a pivot and search the max confidence edge to view 2, the edge A-B with the highest 0.4 score is selected. Then nodes A and B are divided into the sub-graph as indicated by the purple dashed area in the figure, and the uniqueness conflict has been resolved in view 2. 
After that, we detach the sub-graph from the original graph in view 1 and view 2 by cutting off the connections A-C and B-C, represented as transparent dashed arrows in the figure. 
Second, we search the maximum spanning node between the detached sub-graph and nodes in view 3. There are three candidates B-D with a similarity score of 0.37, A-D with 0.35, B-E with 0.36, and the max one is B-D with 0.37. 
So, we merge node D into the sub-graph and cut off all the conflicted edges to solve the uniqueness conflict in view 3.
Here, a maximum spanning subgraph A-B-D is divided from the original graph. 
Third, we repeat the flow as the above two steps layer by layer: choosing a pivot in the remaining nodes and dividing the maximum spanning graph in sequence. 
The flow won't stop until there is no uniqueness conflict.
The pseudo code of the above algorithm is shown in Algorithm~\ref{alg:split_complete_graph}.
When there is no conflict, the remaining nodes will be divided into different subgraphs depending on their connection relations.

Overall, we consider both the implicit connection relations for cycle consistency constraint and the hierarchical maximum spanning for uniqueness constraint.  


\begin{figure*}[htpb!] 
	\centering
	\includegraphics[width=1.0\linewidth]
	{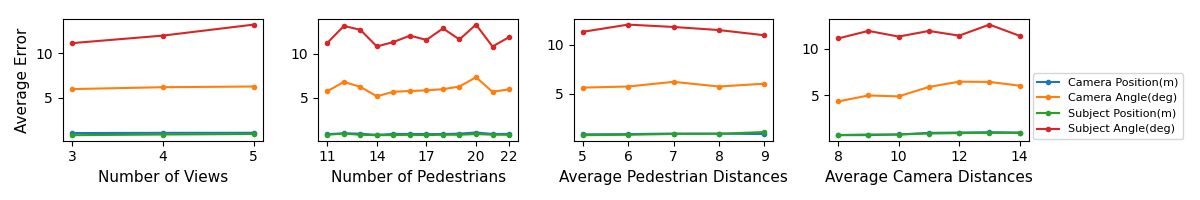} \vspace{-15pt}
	\caption{Results for sensitive analysis.}
	\label{fig:sa}
	\vspace{-5pt}
\end{figure*}

\subsection{Dataset Statistics}
The dataset statistics for CSRD-II, CSRD-V, and CSRD-R are shown in Table~\ref{dataset statistics}.

\begin{table}[htbp!]
	\tabcolsep=0.07cm
	\caption{Dataset statistics.}
	\vspace{-15pt}
	\label{dataset statistics}
	\centering
	\footnotesize
	\begin{spacing}{1.25}
		\begin{center}
			\begin{tabular}{c|cccccc}
				\hline
				& \# Images & \# Annotations & \# Views & \# Sub./Frm. & \# Scenarios \\ \hline
				CSRD-II & 3K     & 51K & 2 & 5-25  & 1       \\ \hline
				CSRD-V  & 5K     & 97 K & 5 & 5-25   & 1       \\ \hline
				CSRD-R  & 15 K    & 170 K & 2-4 & 7-12 & 5   \\ \hline
			\end{tabular}
		\end{center}
	\end{spacing}
	\vspace{-30pt}
\end{table}
\subsection{Details of Comparison Methods}
\label{sec:experiment}
We first compare our method with other methods for the \textit{camera registration} task.\\
$ \bullet $ \textit{DMHA}: DMHA~\cite{han2022connecting_cvpr} achieves the task of camera registration by using the real BEV image. Besides the FPV images, we additionally provide the corresponding BEV image generated by our data engine to DMHA. To evaluate the results, we use the ground-truth position of camera wearers and predicted camera wearers in the generated BEV to calculate the distance and angle errors.\\
$ \bullet $ \textit{SIFT + KNN} and \textit{other deep-learning-based methods}: We also compare with some key point matching based methods, including both the traditional method like SIFT\cite{lowe2004distinctivesift} and the latest CNN based matching methods~\cite{sun2021loftr,sarlin2020superglue,lee2022correlation,zhu2023r2former}. 
The input of both methods is a pair of FPV images and then we can get some key point matching pairs. After that, we use the classical camera pose estimation method with the matched key points to generate the essential matrix and convert it to the relative camera location and (yaw-axis) direction. 
Note that, the error of SIFT is relatively large, some camera position estimation is out of the scene border, in this case, we crop the position of estimation to the outer boundary of that axis. But the same problem does not occur in the deep-learning-based methods for their relatively higher precisions.

For the second task of \textit{subject registration}, {we first compared with a single-view human depth estimation method namely Monoloco++~\cite{monolocopp}.}
Also, we include several  works\cite{multiview_hou1_hou2020multiview,multiview_hou2021multiview} for multi-view detection, which both require the camera calibration to project all views into a shared plane to create the occupancy map. 
\\
$ \bullet $ \textit{Monoloco++}~\cite{monolocopp}: {Monoloco++ is a network trained on KITTI and nuScenes datasets, which is used to predict the 3D-localization and face orientation of each person in the view. We concatenate it with our proposed geometric transformation and subject fusion methods for evaluation.}\\
$ \bullet $ \textit{MVDet} and \textit{MVDetr} \cite{multiview_hou1_hou2020multiview,multiview_hou2021multiview}: These two methods need camera calibrations for generating the results of subject registration. So we calculate the camera calibrations by using the 3D localization of feet (with height = $0$) and the 2D position of the bottom of the bounding box of each person predicted in our methods. 
With the calibration, these two methods generate multi-view human detection predictions (without human identifications) in the BEV.
Then we evaluate the results by using the Hungarian matching algorithm to match the identification of all the predicted points with the ground-truth ones through the minimum spatial distance.

\subsection{Sensitivity Analysis}
Here, we provide the sensitivity analysis of our method to the number of views/pedestrians and the locations of pedestrians and cameras in the figure above.  As shown in Figure~\ref{fig:sa}, we can see that the angle prediction results are more sensitive, but the overall fluctuation of the angle prediction basically stays within 2 degrees, while the position prediction results are quite stable within a very small range.

\subsection{Time Complexity Analysis}

As it is shown in Table~\ref{tab:fps}, we compute the time efficiency of the proposed method. Specifically, we counted the average speed (fps) of different modules and the overall speed.
We can first see that the overall speed is \textit{over the real-time efficiency}.
Moreover, the feature extraction operations in VTM and Association modules take up the main time cost, which can be parallel implemented for multi-view input for further acceleration.

\begin{table}[htbp!]
	\tabcolsep = 0.05cm
	\caption{Time efficiency of different components in our method.}
	\label{tab:fps}
	\vspace{-10pt}
	\centering
	\footnotesize
	\begin{spacing}{1.25}
		\begin{center}
			\begin{tabular}{c|cccc|c}
				\hline
				Module &VTM &Association  & SAM  & Registration   & Overall         \\ \hline
				FPS &126.14  &51.21 & 1490.36 & 3423.51      & 35.19\\
				\hline
			\end{tabular}
		\end{center}
	\end{spacing} 
	\vspace{-0.6cm}
\end{table}

\subsection{Evaluation Metric for Real-world Dataset}

\begin{figure}[htbp!] \vspace{-10pt}
	\centering
	\includegraphics[width=0.55\linewidth]{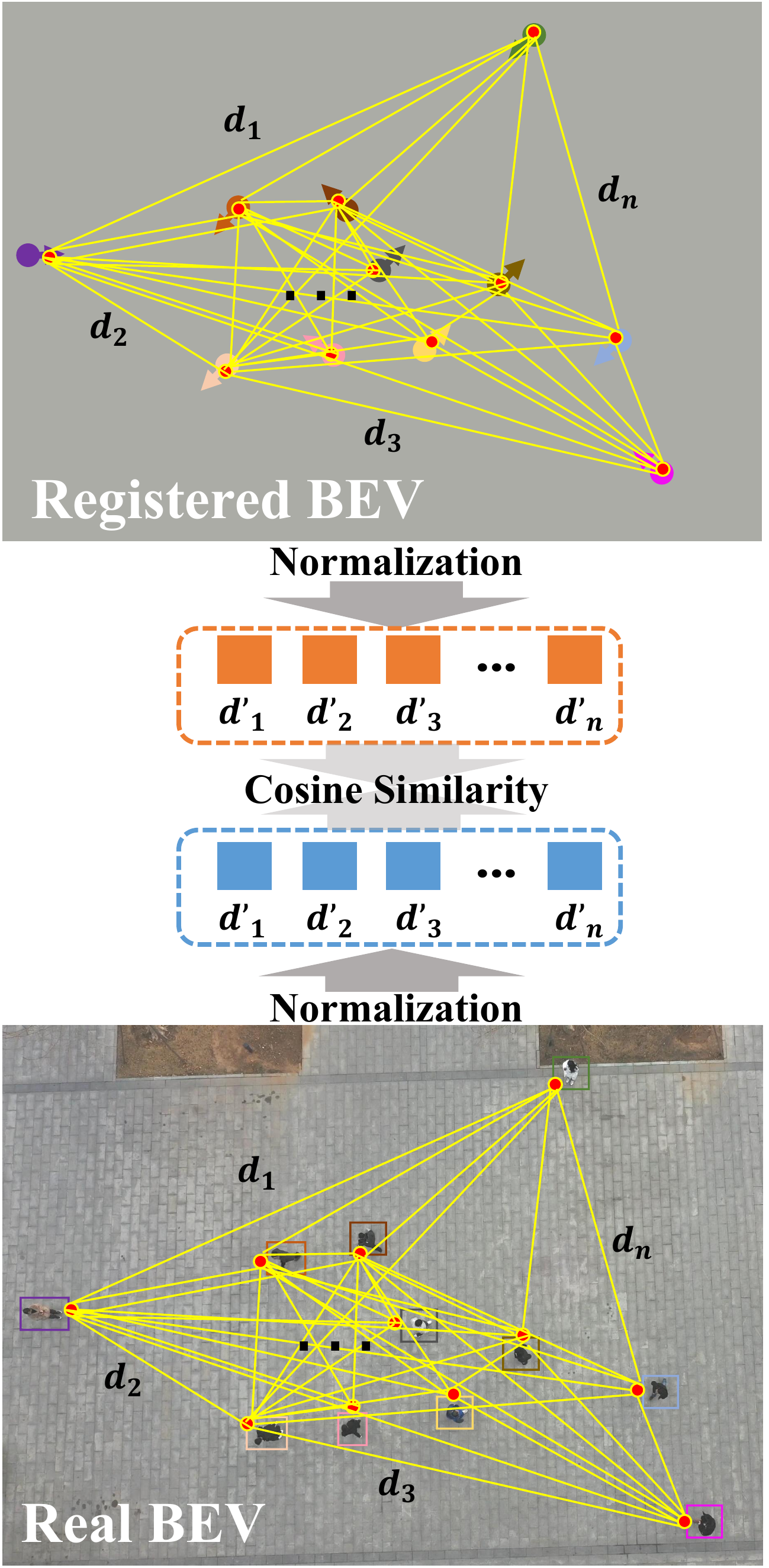} 
	\caption{An illustration of the proposed geometric similarity-based localization metric. We use $d_{i}$ to represent the distance between a pair of subjects in BEV and $d'_{i}$ to represent the normalized distance.}
	\label{fig: cos similarity}
	\vspace{-10pt}
\end{figure}

\textit{Geometric similarity-based localization metric:} 
Evaluating results on real datasets can be challenging for the absence scale between the real BEV and the registered BEV. Here, we propose a geometric similarity-based localization metric, allowing cross-domain performance evaluation between the real BEV and the registered BEV. To achieve this, we first calculate the normalized distances among all subjects in the real and the registered BEVs separately. Then, we flatten the normalized distances as vectors by aligned IDs between the real BEV and the registered BEV. We calculate the cosine similarity between these two vectors and use it to measure the result of cross-domain registration, as shown in Figure~\ref{fig: cos similarity}. 

\subsection{More Visualization Results}

We show more visualization results in different situations. As shown in the following Figures~\ref{fig:great_case3}-\ref{fig:great_case6}, we can see that our method can accomplish the task very well, even in some difficult and special cases.

\begin{figure}[htpb!] 
	\centering
	\includegraphics[width=1.0\linewidth]{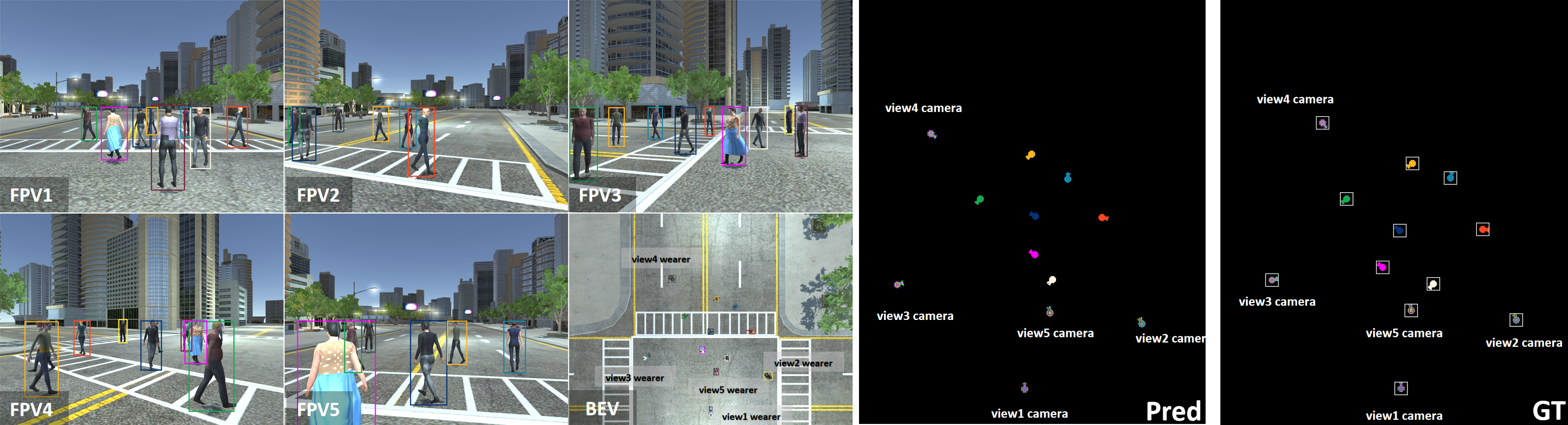} 
	\caption{In this case, the camera wearer of view3 does not appear in any FPV. Our camera registration method can still predict it accurately.}
	\label{fig:great_case3}
		\vspace{-10pt}
\end{figure}

\begin{figure}[htpb!] 
	\centering
	\includegraphics[width=1.0\linewidth]{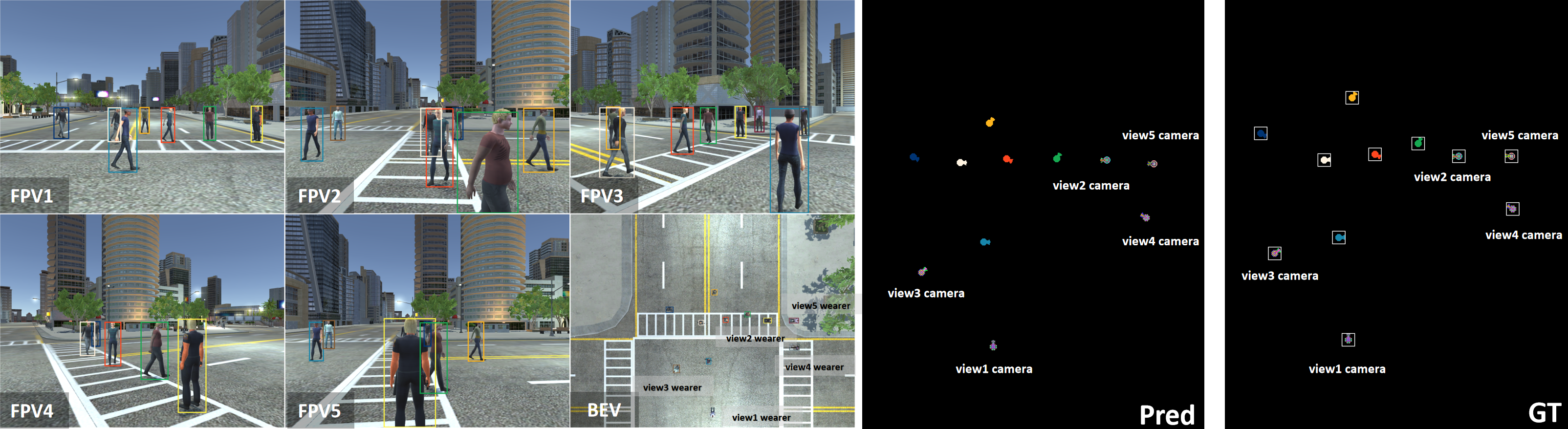} 
	\caption{In this case, six people are standing in a row with serious occlusion. Our method makes full use of information from complementary views to finish the task of registration.}
	\label{fig:great_case4}
		\vspace{-10pt}
\end{figure}

\begin{figure}[htpb!] 
	\centering
	\includegraphics[width=1.0\linewidth]{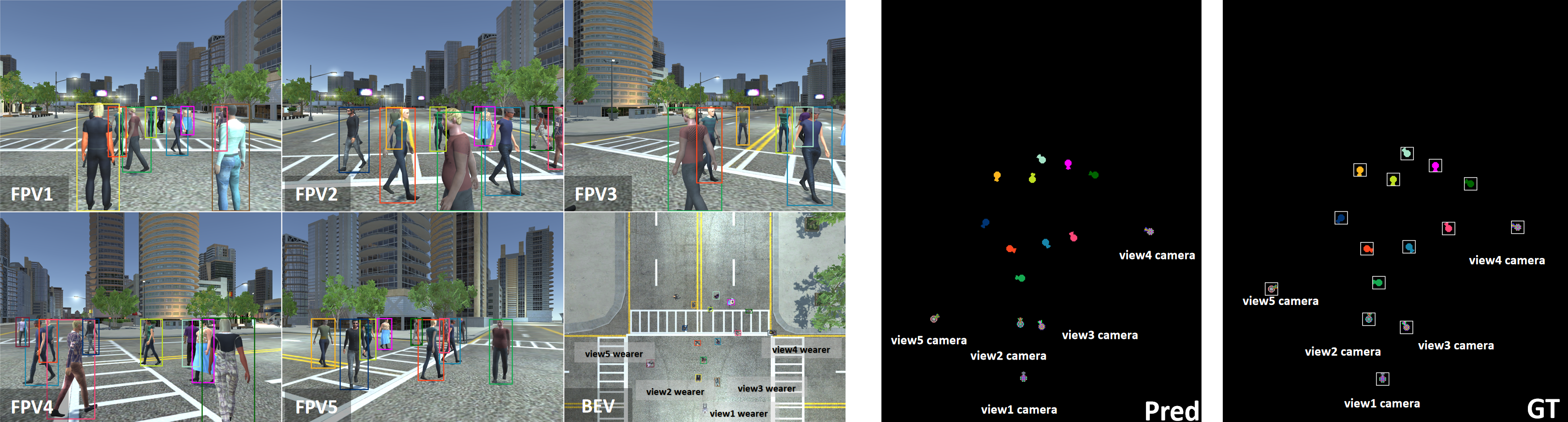} 
	\caption{This case is a dense crowd scene, and the camera is located very close to the crowd, and some of the camera wearers are part of the crowd.}
	\label{fig:great_case5}
\end{figure}

\clearpage

\begin{figure}[htpb!] 
	\centering
	\includegraphics[width=1.0\linewidth]{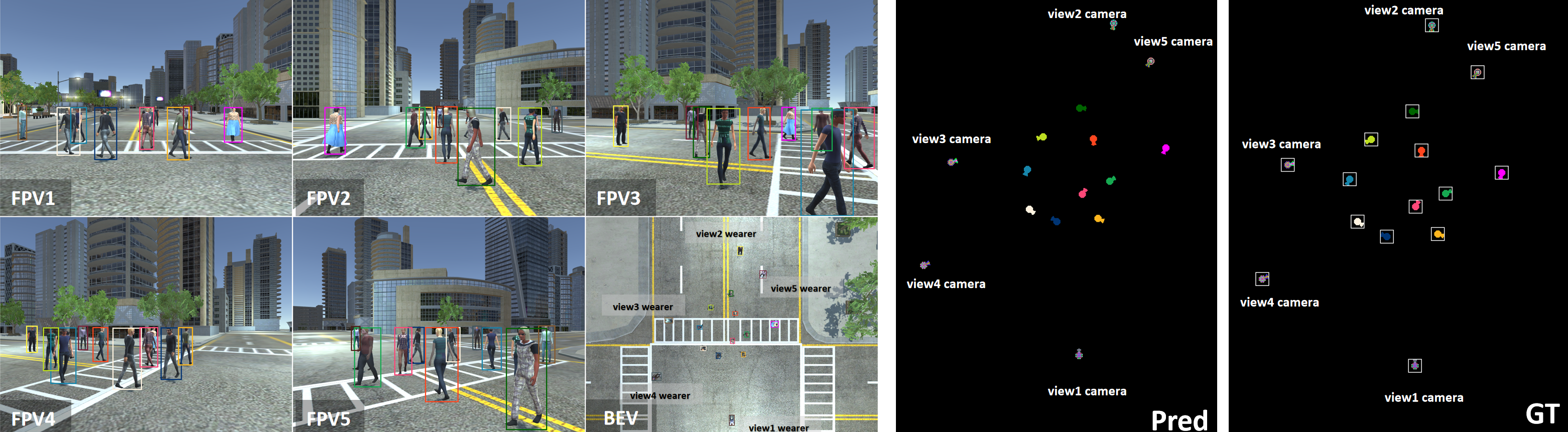} 
	\caption{In this case, camera wearer2 is standing opposite to camera wearer1 and camera wearer4 is standing opposite to camera wearer5, which is the most difficult case of the camera registration task.}
	\label{fig:great_case6}
\end{figure}

{
    \small
    \bibliographystyle{ieeenat_fullname}
    \bibliography{main}
}


\end{document}